\documentclass[a4]{article}

\usepackage{hyperref}
\usepackage{subcaption}
\usepackage{amsmath}
\usepackage{amsfonts}
\usepackage{authblk}
\usepackage{graphicx}
\usepackage{booktabs}
\usepackage{multirow}

\newcommand*{\QED}[1][$\square$]{%
	\leavevmode\unskip\penalty9999 \hbox{}\nobreak\hfill
	\quad\hbox{#1}%
}

\begin{document}

\title{\textbf{Epistemic Uncertainty Quantification in Deep Learning Classification by the Delta Method}}
\author[1,2]{Geir K. Nilsen}
\author[1]{Antonella Z. Munthe-Kaas}
\author[1]{Hans J. Skaug}
\author[1]{Morten Brun}
\affil[1]{Department of Mathematics, University of Bergen}
\affil[2]{geir.kjetil.nilsen@gmail.com}
\date{}                     
\setcounter{Maxaffil}{0}
\renewcommand\Affilfont{\itshape\small}





\maketitle

\begin{abstract}
The Delta method is a classical procedure for quantifying epistemic uncertainty in statistical models, but its direct application to deep neural networks is prevented by the large number of parameters $P$. We propose a low cost variant of the Delta method applicable to $L_2$-regularized deep neural networks based on the top $K$ eigenpairs of the Fisher information matrix. We address efficient computation of full-rank approximate eigendecompositions in terms of either the exact inverse Hessian, the inverse outer-products of gradients approximation or the so-called Sandwich estimator. Moreover, we provide a bound on the  approximation error for the uncertainty of the predictive class probabilities. We observe that when the smallest eigenvalue of the Fisher information matrix is near the $L_2$-regularization rate, the approximation error is close to zero even when $K\ll P$. A demonstration of the methodology is presented using a TensorFlow implementation, and we show that meaningful rankings of images based on predictive uncertainty can be obtained for two LeNet-based neural networks using the MNIST and CIFAR-10 datasets. Further, we observe that false positives have on average a higher predictive epistemic uncertainty than true positives. This suggests that there is supplementing information in the uncertainty measure not captured by the classification alone.

\end{abstract}

\section{Introduction}
The predictive probabilities at the output layer of neural network classifiers are often misinterpreted as model (epistemic) uncertainty \cite{gal}. Bayesian statistics provides a coherent framework for representing uncertainty in neural networks \cite{mackay, goodfellowetal}, but has not so far gained widespread use in deep learning -- presumably due to the high computational cost that traditionally comes with second-order methods. Recently, \cite{gal} developed a theoretical framework which casts dropout at test time in deep neural networks as approximate Bayesian inference. Due to its mathematical elegance and negligible computational cost, this work has caught great interest in a variety of different fields \cite{loquercio, litjens, zhu}, but has also generated questions as to what types of uncertainty these approximations actually lead \cite{osband, osband2} and what types are relevant \cite{gal2}. For a general treatment of uncertainty in machine learning, we refer to \cite{hullermeier}.

Epistemic uncertainty is commonly understood as the reducible component of uncertainty -- the uncertainty of the model itself, or its parameters. In our context this amounts to the uncertainty in the estimated class probabilities due to limited amount of training data.	While the epistemic uncertainty can be reduced by increasing the amount of training data, the other component of uncertainty known as aleatoric uncertainty, is irreducible and stems from the uncertainty in the label assignment process \cite{song}. However, in this paper we only address the epistemic part, and treat the labels as constant when estimating uncertainty.

Our approach goes back to the work of \cite{mackay}, and we show that the above reasoning leads to the method known as the Delta method\footnote{Also known as the Laplace approximation.} \cite{hoef,mcfadden,khosravi} in statistics. However, as the Delta method depends on the empirical Fisher information matrix which grows quadratically with the number of neural network parameters $P$ -- its direct application in modern deep learning is prohibitively expensive. We therefore propose a low cost variant of the Delta method applicable to $L_2$-regularized deep neural networks based on the top $K$ eigenpairs of the Fisher information matrix. We address efficient computation of full-rank approximate eigendecompositions in terms of either the exact inverse Hessian, the inverse outer-products of gradients (OPG) approximation or the so-called Sandwich estimator. Further, we exhibit the fact that deep learning classifiers tend to be heavily over-parameterized. This leads to flat Fisher information eigenvalue spectra which we show can be exploited in terms of a simple linearization.

The theoretical Fisher information matrix is always positive (semi)-definite, and we constrain our empirical counterpart to be the same. Recent research \cite{sagun, sagun2, alain, ghorbani}, consistent with our own observations, show that the exact Hessian after training is rarely positive definite in deep learning. To mitigate this, we propose a simple correction of the right tail of the Hessian eigenvalue spectrum to achieve positive definiteness. We corroborate our choice with two observations: a) negative eigenvalues of the Hessian matrix are highly stochastic across different weight initialization values, and b) correcting the eigenvalue spectrum to achieve positive definiteness yields stable predictive epistemic uncertainty estimates which are perfectly correlated with the estimates based on the OPG approximation -- which by construction is always positive (semi)-definite \cite{martens}.

As the computational cost of the exact inverse Hessian matrix or its full eigendecomposition is prohibitively expensive in deep learning, we propose to use the Lanczos iteration \cite{trefethen} in combination with Pearlmutter's technique \cite{pearlmutter} to compute the needed eigenpairs. Consequently, the matrix inversion will be straightforward, and the net computational complexity will be $O(SPN)$ time and $O(KP)$ space, where $N$ is the number of training examples and $S$ is the number of Lanczos-Pearlmutter steps required to compute $K$ eigenpairs.

Also the inverse OPG approximation or its full eigendecomposition is prohibitively costly in deep learning. Even if we disregard the inversion and the quadratic space complexity, one is first left to compute and store the $N \times P$-dimensional Jacobian matrix. In deep learning software provisions based on backward-mode automatic differentiation, only the sum of mini-batch gradients can be computed efficiently. We therefore propose to compute mini-batches of the Jacobian using efficient per-example gradients \cite{nilsen} in combination with incremental singular value decompositions \cite{levy}. Since the OPG approximation can be written as a Jacobian matrix product, its eigenvectors will be the right singular vectors of the Jacobian, and its eigenvalues the squared singular values. This leads to a computational complexity of $O(KPN)$ time and $O(KP)$ space, also accounting for the inversion. The Sandwich estimator requires both the inverse Hessian and the OPG approximation, and is thus $O(max\{K,S\}PN)$ time and $O(KP)$ space.

This work is a continuation of \cite{nilsen}, and we here introduce the fully deterministic \cite{nagarajan} open sourced TensorFlow module \texttt{pydeepdelta} \cite{pydeepdeltamodule}, and illustrate the methodology on two LeNet-based convolutional neural network classifiers using the MNIST and CIFAR-10 datasets.

The paper is organized as follows: In Section \ref{section:definitions} we give definitions which will be used throughout the paper. In Section \ref{section:classicaldeltamethod} we review the Delta method in a deep learning classification context, and in Section  \ref{section:deepdeltamethod} we outline the details of the proposed methodology. In Section \ref{sec:demo} we demonstrate the method, and finally, in Section \ref{sec:summary} we summarize the paper and give some concluding remarks and ideas of future work.

\section{Deep Neural Networks}\label{section:definitions}
We use a feed-forward neural network architecture with dense layers to introduce terminology and symbols, but emphasize that the theory presented in the paper is directly applicable to any $L_2$-regularized architecture.

\subsection{Architectural}
\begin{figure*}[h]
	\centering
	\includegraphics[scale=0.36]{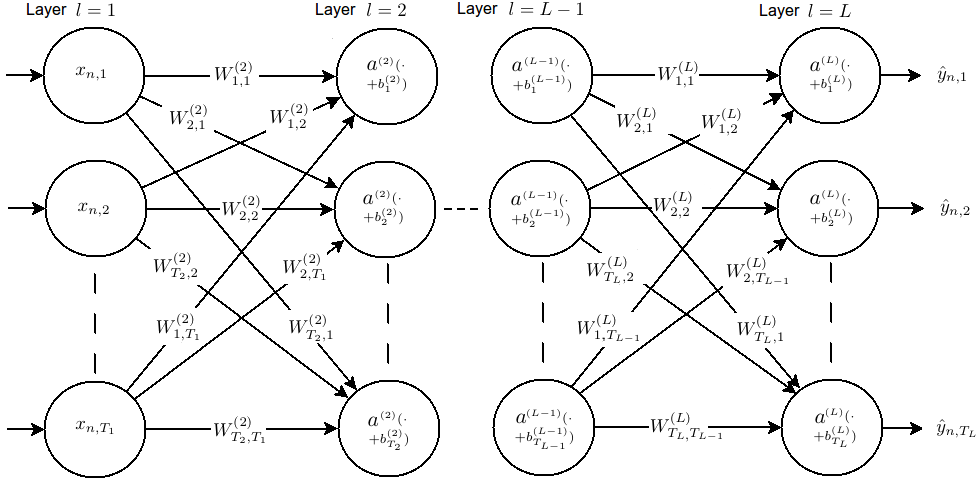}
	\caption{A feed-forward neural network with dense layers.}
	\label{fig:ffnn}
\end{figure*}

A feed-forward neural network is shown in Figure \ref{fig:ffnn}. There are $L$ layers $l=1,2,...,L$ with $T_l$ neurons in each layer. The input layer $l=1$, is represented by the input vector $x_n = \begin{pmatrix}x_{n,1} & x_{n,2} & \hdots & x_{n,T_1}\end{pmatrix}^T$ where $n=1,2,...,N$ is the input index. Furthermore, there are $L-2$ dense hidden layers, $l=2,3,...,L-1$, and a dense output layer $l=L$, each represented by weight matrices $W^{(l-1)} \in \mathbb{R}^{T_{l} \times T_{l-1}}$, bias vectors $b^{(l)} \in {\mathbb{R}^{T_l}}$ and vectorized activation functions $a^{(l)}$.

\subsection{Parameter Vectors}\label{sec:paramvecs}
The total number of parameters in the model shown in Figure \ref{fig:ffnn} can be written,
\begin{equation}
	P = \sum_{l=2}^{L} P^{(l)} = \sum_{l=2}^{L} T_{l-1} T_l + T_l,
\end{equation}
where $P^{(l)}$ denotes the number of parameters in layer $l$. By definition, $P^{(1)}=0$ since the input layer contains no weights or biases. Furthermore, we define parameter vectors representing the layer-wise weights and biases as follows,
\begin{align}
	\omega^{(l)} &= \begin{bmatrix}\text{vec}(W^{(l)}) \\ b^{(l)}\end{bmatrix} \in \mathbb{R}^{P^{(l)}},
	\label{eq:flatten}
\end{align}
for $l=2,3,\hdots,L$, with components $\omega_i^{(l)}$, $i=P^{(l-1)}+1, P^{(l-1)}+2, \hdots, P^{(l)}$. The notation $\text{vec}(W)$ denotes a row-wise vectorization\footnote{Standard method in TensorFlow: tf.reshape(W, [-1])} of the matrix $W^{A \times B}$ into a column vector of dimension $\mathbb{R}^{AB}$. In the rest of the paper, we consider the full model and define the parameter vector,
\begin{equation}
	\omega = \begin{bmatrix} \omega^{(2)} \\ \omega^{(3)} \\ \vdots \\ \omega^{(L)}\end{bmatrix} \in \mathbb{R}^{P}.\label{eq:paramvec}
\end{equation}
\subsection{Training, Model and Cost Function}
The \textit{model function} $f: \mathbb{R}^{T_1\times P} \rightarrow \mathbb{R}^{T_L}$ associated to the architecture shown in Figure \ref{fig:ffnn} is defined as
\begin{equation}\label{eq:modelfunc}
	f(x_n, \omega) = a^{(L)}[W^{(L)}a^{(L-1)}(\cdots a^{(2)}\lbrace W^{(2)}x_n + b^{(2)}\rbrace + \cdots ) + b^{(L)}].
\end{equation}
We use a softmax cross-entropy \textit{cost function} $C: \mathbb{R}^{P}\rightarrow \mathbb{R}$ and require $L_2$-regularization with a rate factor $\lambda > 0$,
\begin{align}
	C(\omega) &= \frac{1}{N}\sum_{n=1}^{N}C_n(y_{n}, \hat{y}_n) + \frac{\lambda}{2}\sum_{p=1}^P\omega_p^2\nonumber\\
			  &= \frac{1}{N}\sum_{n=1}^N\left( - \sum_{m=1}^{T_L} y_{n,m} \text{log } \hat{y}_{n,m}\right) + \frac{\lambda}{2}\sum_{p=1}^P\omega_p^2,\label{eq:costfunction}
\end{align}
where $y_{n}$ represents the target vector for the $n$th example ($N$ examples), and where $\hat{y}_{n} = f(x_n, \omega)$ represents the corresponding prediction vector obtained by evaluating the \textit{model function} (\ref{eq:modelfunc}) using the input vector $x_n$ and the parameter vector (\ref{eq:paramvec}). The activation function $a^{(L)}: \mathbb{R}^{T_L}\rightarrow\mathbb{R}^{T_L}$ in the output layer is the vectorized softmax function defined as
\begin{align}
	a^{(L)}(z) &= \text{softmax}(z)\nonumber\\ &= \frac{\text{exp}{(z)}}{\sum_{m=1}^{T_L}\text{exp}{(z_{m})}},
\end{align}
where $\text{exp}(\cdot)$ denotes the vectorized exponential function. 
Training of the neural network can be defined as finding an `optimal' parameter vector $\hat{\omega}$ by minimizing the cost function (\ref{eq:costfunction}),
\begin{equation}
	\hat{\omega} = \underset{\omega \in \mathbb{R}^{P}}{\text{arg min}~C(\omega)}.\label{eq:omegahat}
\end{equation}
\section{The Delta Method}\label{section:classicaldeltamethod}
The Delta method \cite{hoef} views a modern deep neural network as a (huge) non-linear regression. In our classification setting, we regard the labels as constant, and thus the epistemic component of the uncertainty associated with predictions of an arbitrary input example $x_0$ reduces to the evaluation of the covariance matrix of the network outputs \cite{khosravi}. By a first-order Taylor expansion \cite{grosse}, it can be shown that the covariance matrix of the network outputs $\hat{y}_0$, i.e.\ the model function (\ref{eq:modelfunc}), can be approximated by
\begin{equation}
	\text{Cov}(\hat{y}_0) \approx F\Sigma F^T \in \mathbb{R}^{T_L\times T_L}\label{eq:taylor},
\end{equation}
where
\begin{equation}\label{eq:jacobian}
	F = \begin{bmatrix}F_{ij}\end{bmatrix}\in \mathbb{R}^{T_L\times P},~F_{ij} = \frac{\partial}{\partial \omega_j} f_i(x_0, \omega)\bigg\rvert_{\omega=\hat{\omega}}
\end{equation} 
is the Jacobian matrix of the model function, and where $\Sigma$ is the covariance matrix of the model parameter estimate $\hat{\omega}$. For a given $x_0$, an approximate standard deviation of $\hat{y}_0$ is provided by the formula
\begin{equation}
	\sigma(x_0) \approx \sqrt{\text{diag}\big(F\Sigma F^T\big)}\in \mathbb{R}^{T_L}\label{eq:sigma1}.
\end{equation}
Equation (\ref{eq:sigma1}) means that when the neural network predicts for an input $x_0$, the associated epistemic uncertainty per class output is determined by a linear combination of parameter sensitivity (e.g.\ $F$) and parameter uncertainty (e.g.\ $\Sigma$). Parameter sensitivity ($F$) prescribes the amount of change in the neural network output for an infinitesimal change in the parameter estimates, whereas the parameter uncertainty ($\Sigma$) prescribes the amount of uncertainty in the parameter estimates themselves.

We apply and compare three different approximations to $\Sigma$. The first one is called the \textbf{Hessian estimator}, and is defined by
\begin{equation}
	\Sigma^{\text{H}} = \frac{1}{N}H^{-1} = \frac{1}{N}\left[\frac{1}{N}\sum_{n=1}^N\frac{\partial^2 C_n}{\partial \omega \partial \omega^T}\bigg\rvert_{\omega=\hat{\omega}} + \lambda I\right]^{-1}\in \mathbb{R}^{P\times P},\label{eq:hess}
\end{equation}
where $H$ is the empirical Hessian matrix of the cost function evaluated at~$\hat{\omega}$.

The second estimator is called the \textbf{Outer-Products of Gradients (OPG) estimator} and is defined by
\begin{equation}
	\Sigma^{\text{G}} = \frac{1}{N}G^{-1} = \frac{1}{N}\left[\frac{1}{N}\sum_{n=1}^N \frac{\partial C_n}{\partial \omega} \frac{\partial C_n}{\partial \omega}^T\bigg\rvert_{\omega=\hat{\omega}} + \lambda I\right]^{-1}\in \mathbb{R}^{P\times P},\label{eq:opg}
\end{equation}
where the summation part of $G$ corresponds to the empirical covariance of the gradients of the cost function evaluated at $\hat{\omega}$. Finally, the third estimator is known as the \textbf{Sandwich estimator} \cite{freedman, schulam} and is defined by
\begin{equation}
	\Sigma^{\text{S}} = \frac{1}{N}H^{-1}GH^{-1}\in \mathbb{R}^{P\times P}. \label{eq:sandwich}
\end{equation}

Across various fields and contexts, the two famous equations \eqref{eq:hess} and \eqref{eq:opg} are often presented and interpreted differently, and the inconsistency in the vast literature is nothing but intriguing. We therefore feel that their appearance in this paper requires some elaboration. Firstly, for the Hessian estimator \eqref{eq:hess}, we note that the differentials act only on the data dependent part of the cost function \eqref{eq:costfunction}, $C_n$, so the second term, $\lambda I$, here comes from the second-order derivatives of the $L_2$-regularization term. Secondly, for the OPG estimator \eqref{eq:opg}, also here the differentials act on the data dependent part of the cost function, but the crucial detail often confused or let out in the literature comes with the second term, $\lambda I$: under $L_2$-regularization it must be added explicitly in order for $G$ to be asymptotically equal to $H$ (See the Appendix \ref{sec:appendix} for a proof) -- as is the primary motivation of the OPG estimator as a plug-in replacement of the Hessian estimator in the first place. If let out, $G$ will almost always be singular \cite{watanabe, murfet}, and thus cannot be used in \eqref{eq:opg}.

At this point, we can see that two fundamental difficulties arise when applying the Delta method in deep learning: a) the sheer size of the covariance matrix grows quadratically with $P$, and 2) the covariance matrix must be positive definite. In other words, we are virtually forced to compute and store the full covariance matrix, and are in terms of the Hessian estimator dependent on that the optimizer can find a true local (or global) minimum of the cost function. Nevertheless, with the OPG and the Sandwich estimators, the second obstacle is virtually inapplicable since they by definition always will be positive definite when~$\lambda>0$.

In the next section we present methodology that addresses both these aspects. We present an indirect correction leaving the Hessian estimator positive definite, and introduce methodology with computational time and space complexity which is linear in~$P$.

\section{The Delta Method in Deep Learning}\label{section:deepdeltamethod}
\begin{figure*}[h]
	\centering
	\includegraphics[scale=0.5]{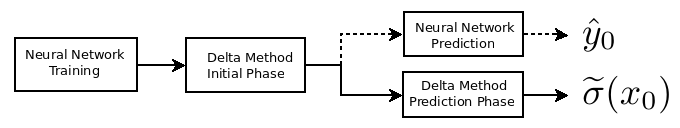}
	\caption{The Delta method for quantifying the predictive epistemic uncertainty $\widetilde{\sigma}(x_0)$ of $\hat{y}_0 = f(x_0, \hat{w})$ in deep learning (solid line).}
	\label{fig:ddflowchart}
\end{figure*}

We present our approach to the Delta method in deep learning as a procedure carried out in two phases after the neural network has been trained. See Figure \ref{fig:ddflowchart}.

The first phase -- the `initial phase' -- is carried out only once, with the scope of indirectly computing full-rank, positive definite approximations of the covariance matrices \eqref{eq:hess}, \eqref{eq:opg} or \eqref{eq:sandwich} based on approximate eigendecompositions of $H$ and $G$. The second phase -- the `prediction phase' -- is carried out hand in hand with the regular neural network prediction process (\ref{eq:modelfunc}), and is used to approximate the epistemic component of the predictive uncertainty governed by (\ref{eq:sigma1}) using the indirect covariance matrix approximation found in the `initial phase'.

In the next sections, we address the following aspects of the proposed methods: a) how to efficiently compute eigenvalues and eigenvectors of the Hessian estimator via the Lanczos iteration and exact Hessian vector products, b) how to efficiently compute eigenvalues and eigenvectors of the OPG estimator via incremental singular value decompositions, c) how to combine the former two to obtain an approximation of the Sandwich estimator, and d) how to apply these estimators to efficiently compute an approximation of (\ref{eq:sigma1}).

\subsection{Computing Eigenvalues and Eigenvectors of the Covariance Matrix}\label{sec:lanczos}
The full eigendecomposition of the covariance matrix in \eqref{eq:sigma1} is defined by
\begin{equation}
	\Sigma = Q\Lambda^{-1} Q^T\label{eq:eigendecomp} \in \mathbb{R}^{P \times P},
\end{equation}
where $Q \in \mathbb{R}^{P \times P}$ is the matrix whose $k$th column is the eigenvector $q_k$ of $\Sigma$, and $\Lambda \in \mathbb{R}^{P \times P}$ is the diagonal matrix whose elements are the corresponding eigenvalues, $\Lambda_{kk} = \lambda_k$. We assume that the eigenvalues are algebraically sorted so that $\lambda_1 \ge \lambda_2 \ge \hdots \ge \lambda_P$. Note that in terms of the Hessian estimator, the eigenvalues are precisely the second derivatives of the cost function along the principal axes of the ellipsoids of equal cost, and that $Q$ is a rotation matrix which defines the directions of these principal axes \cite{lecun2}.

For the Hessian estimator \eqref{eq:hess}, the Lanczos iteration \cite{trefethen} can be applied to find $K < P$ eigenvalues (and corresponding eigenvectors) in $O(SNP)$ time and $O(KP)$ space when Pearlmutter's technique \cite{pearlmutter} is applied inside the iteration \cite{nilsen}. Pearlmutter's technique can simply be described as a procedure based on two-pass back-propagations of complexity $O(NP)$ time and $O(P)$ space to obtain exact Hessian vector products without requiring to keep the full Hessian matrix in memory. The number $S$ denotes the number of Lanczos iterations to reach convergence. We observe that the convergence of the Lanczos algorithm is quite fast in our experiments, and we find that $S$ is practically orders of magnitude less than $P$.

For the OPG estimator \eqref{eq:opg}, a slightly different approach can be applied. Since the OPG estimator can be written as a Jacobian matrix product \cite{nilsen}, we get by the singular value decomposition that its eigenvectors will be the right singular vectors of the Jacobian, and its eigenvalues the squared singular values. Mini-batches of the Jacobian matrix can easily be obtained by standard back-propagation, and so an incremental singular value decomposition \cite{levy, cardot} can be applied to each mini-batch. The computational cost is thus $O(KNP)$ time and $O(KP)$ space. The Sandwich estimator combines the Hessian and the OPG approximation via the product \eqref{eq:sandwich}, and thus has a computational complexity of $O(max\{K,S\}NP)$ time and $O(KP)$ space. The computational complexity of the outlined methodology is summarized in Table \ref{tab:runtimes}\footnote{Assuming naive matrix multiplication}.

\begin{table}[h]
	\scalebox{0.8}{
		\begin{tabular}{c|c|c|c|c|}
			\cline{2-5}
			\multicolumn{1}{l|}{}                   & \multicolumn{2}{c|}{\textbf{Initial Phase}} & \multicolumn{2}{c|}{\textbf{\begin{tabular}[c]{@{}c@{}}Prediction Phase \\ (Per-Example)\end{tabular}}} \\ \cline{2-5} 
			\multicolumn{1}{l|}{}                   & \textbf{Time}    & \textbf{Space}           & \textbf{Time}                                              & \textbf{Space}                             \\ \hline
			\multicolumn{1}{|c|}{\textbf{Hessian}}  & $O(SPN)$         & \multirow{3}{*}{$O(KP)$} & \multirow{3}{*}{$O(T_LPK + T_L^2K + K^2T_L)$}              & \multirow{3}{*}{$O(max\{K,T_L\}P)$}             \\ \cline{1-2}
			\multicolumn{1}{|c|}{\textbf{OPG}}      & $O(KPN)$         &                          &                                                            &                                            \\ \cline{1-2}
			\multicolumn{1}{|c|}{\textbf{Sandwich}} & $O(max\{K,S\}PN)$  &                          &                                                            &                                            \\ \hline
		\end{tabular}
	}
	\caption{The computational complexity of the outlined methodology is linear in $P$ across both phases. }
	\label{tab:runtimes}
\end{table}

Our TensorFlow module \texttt{pydeepdelta} \cite{pydeepdeltamodule} utilizes the Lanczos implementation available in the SciPy distribution \cite{scipy}, as well as the incremental singular value decomposition available in the scikit-learn distribution \cite{sklearn}.

\subsection{The Eigenvalue Spectra of $H$ and $G$}\label{sec:covapprox}
To better understand the proposed covariance approximations, we first need to explore the prototypical deep learning eigenvalue spectrum of the empirical Hessian matrix $H$ \eqref{eq:hess} and the empirical covariance of the gradients $G$ \eqref{eq:opg}. To this end, we introduce two LeNet-based convolutional neural network classifiers using the MNIST and CIFAR-10 datasets, and draw parallels to the findings in the literature.

\subsubsection{Classifier Architectures, Parameters and Training}\label{sec:classifier}
The MNIST classifier has $L=6$ layers, layer $l=1$ is the input layer represented by the input vector. Layer $l=2$ is a $3\times 3\times 1\times 32$ convolutional layer followed by max pooling with stride equal to $2$ and with a ReLU activation function. Layer $l=3$ is a $3\times 3\times 32\times 64$ convolutional layer followed by max pooling with a stride equal to $2$, and with ReLU activation function. Layer $l=4$ is a $3\times 3\times 64\times 64$ convolutional layer with ReLU activation function. Layer $l=5$ is a $576\times 64$ dense layer with ReLU activation function, and the output layer $l=6$ is a $64\times T_L$ dense layer with softmax activation function, where the number of classes (outputs) is $T_L=10$. The total number of parameters is $P=93322$.

The CIFAR-10 classifier has $L=6$ layers, layer $l=1$ is the input layer represented by the input vector. Layer $l=2$ is a $3\times 3\times 3\times 32$ convolutional layer followed by max pooling with stride equal to $2$ and with a ReLU activation function. Layer $l=3$ is a $3\times 3\times 32\times 64$ convolutional layer followed by max pooling with a stride equal to $2$, and with ReLU activation function. Layer $l=4$ is a $3\times 3\times 64\times 64$ convolutional layer with ReLU activation function. Layer $l=5$ is a $1024\times 64$ dense layer with ReLU activation function, and the output layer $l=6$ is a $64\times 10$ dense layer with softmax activation function, where the number of classes (outputs) is $T_L=10$. The total number of parameters is $P=122570$.

We apply random normal weight initialization and zero bias initialization. We use (\ref{eq:costfunction}) as the cost function with a $L_2$-regularization rate $\lambda=0.01$. We utilize the Adam optimizer \cite{kingma, bottou} with a batch size of $100$, and apply no form of randomized data shuffling. To ensure convergence (e.g. $||\nabla C(\hat{\omega})||_2 \approx 0$) we apply the following learning rate schedules given by the following (step, rate) pairs:  MNIST = $\lbrace(0, 10^{-3}), (60\text{k}, 10^{-4}), (70\text{k}, 10^{-5}), (80\text{k}, 10^{-6})\rbrace$ and CIFAR-10 = $\lbrace(0, 10^{-3}), (55\text{k}, 10^{-4}), (85\text{k}, 10^{-5}), (95\text{k}, 10^{-6}, (105\text{k}, 10^{-7})\rbrace$. For MNIST, we stop the training after $90,000$ steps -- corresponding to a training accuracy of $0.979$, test accuracy $0.981$, training cost $C(\hat{\omega}) = 0.257$ and a gradient norm $||\nabla C(\hat{\omega})||_2 = 0.016$. For CIFAR-10, we stop the training after $115000$ steps -- corresponding to a training accuracy of $0.701$, test accuracy $0.687$, training cost $C(\hat{\omega}) = 1.284$ and a gradient norm $||\nabla C(\hat{\omega})||_2 = 0.030$.

\subsubsection{The Eigenvalue Spectrum Approximation}\label{sec:eigenvaluespectrum}
The general assumption in deep learning is that $H$ after training is not positive definite and mostly contain eigenvalues close to zero \cite{sagun, sagun2, alain, ghorbani, granziol, watanabe}. The same holds true for $G$ although it by definition must at least be positive semi-definite \cite{martens}. However, given the discussion in Section \ref{section:classicaldeltamethod}, we know that $L_2$-regularization with rate $\lambda/2$ has the effect of shifting the eigenvalues of $H$ and $G$ upwards by $\lambda$.

\begin{figure}[h]
	\centering
	\begin{subfigure}{0.5\textwidth}
		\includegraphics[scale=0.35]{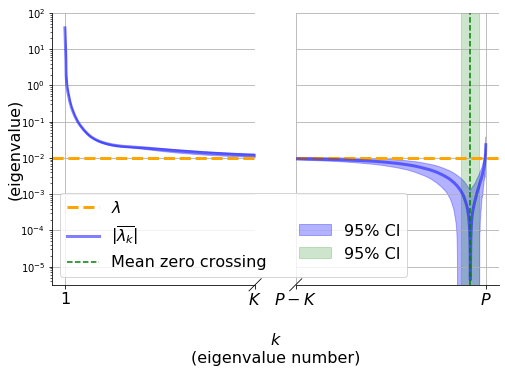}
		\caption{H}
		\label{fig:eigenvaluespectrumH}
	\end{subfigure}%
	\begin{subfigure}{0.5\textwidth}
		\includegraphics[scale=0.35]{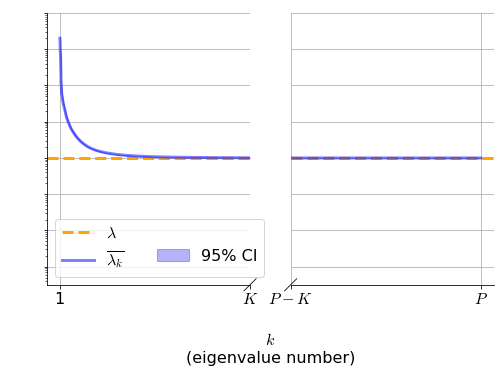}
		\caption{G}
		\label{fig:eigenvaluespectrumG}
	\end{subfigure}
	\caption{Log scale eigenvalue magnitude spectra of $H$ and $G$ showing the $K=1500$ largest (left tail subspace) and the $K=1500$ smallest (right tail subspace) eigenvalues and their variation across sixteen trained instances of the MNIST network distinguished only by a different random weight initialization.}
	\label{fig:eigenvaluespectrum}
\end{figure}

To test this hypothesis, we study the $K=1500$ algebraically largest and the $K=1500$ algebraically smallest eigenvalues of $H$ and $G$ for 16 trained instances of the MNIST network defined in Section \ref{sec:classifier}. These sixteen networks are thus only distinguished from each other by a different random weight initialization prior to training. The two corresponding log-scale eigenvalue magnitude spectra are shown in Figure \ref{fig:eigenvaluespectrum}.

Firstly, we note that in the midpoint gaps of the spectra, there are $P-2K=90195$ `missing' central eigenvalues which we have not computed. Since the eigenvalues are sorted in decreasing order, all the central eigenvalues must be close to the $L_2$-regularization rate $\lambda$. We refer to this part of the eigenvalue spectrum as the gap. Secondly, we note that the confidence intervals in the plots are taken across instance space, thus telling how the eigenvalue spectrum change based on the 16 random weight initializations. In both plots, the blue confidence interval tells that the largest eigenvalues of $H$ and $G$ (called left tail) are stable across the 16 trained networks, but the smallest eigenvalues of $H$ are changing dramatically (called right tail, left plot). On the contrary, all the eigenvalues of $G$ are stable. Thirdly, as shown by the green vertical dotted line in the upper plot representing the mean zero-crossing, $H$ is clearly not positive definite -- even with $L_2$-regularization. The green confidence interval around the zero-crossing shows that the number of negative eigenvalues also change across the networks.

In \cite{granziol} it was hypothesized that negative Hessian eigenvalues are caused by a discrepancy between the empirical Hessian (e.g. $H$) and its theoretical counterpart (expected Hessian) in which the summation of \eqref{eq:hess} is replaced with an expectation so that effectively $N\rightarrow \infty$. They showed that as $N$ grows (holding $\hat{\omega}$ fixed), the empirical right tail grows toward $\lambda$ whereas the rest of the spectrum is stable. Supported by the fact that $H$ and $G$ will be equal in expectation (Appendix \ref{sec:appendix}), the expected Hessian eigenvalue spectrum might be more similar to that of $G$ where all the eigenvalues are greater than equal to $\lambda$. In line with these ideas and the empirical evidence presented in Figure \ref{fig:eigenvaluespectrum}, we assume that all the smallest eigenvalues of $H$ in the right tail are inherently noisy, and should not be used by the Hessian estimator. Therefore, with reference to Figure \ref{fig:gapanalysis}, for the Hessian estimator, we a) calculate all the eigenpairs in the left tail, b) approximate all the eigenvalues in the gap and c) extrapolate the eigenvalues from the gap into the right tail. The eigenvectors corresponding to the gap and right tail can implicitly be accounted for by orthonormality as discussed in the next section.

For the OPG estimator, the same principle applies apart from that the extrapolation inherently becomes a part of the gap subspace approximation because we know that $G$ always is positive definite when $\lambda > 0$. Finally, for the Sandwich estimator, we simply apply the aforementioned procedures and estimate the product \eqref{eq:sandwich}.

\subsection{Closing the Gap}\label{sec:closingthegap}
Based on the observations in the previous section, we now propose a partitioning of the eigendecomposition which reveals that full-rank, positive definite approximations of the Hessian and OPG estimators can be obtained by computing only the eigenpairs corresponding to the $K$ algebraically largest eigenvalues of $H$ and $G$ respectively. Finally, we show how to use these approximations to construct an approximation of the Sandwich estimator.

\subsubsection{The Hessian and OPG Estimators}
\begin{figure}[h]
	\centering
	\includegraphics[scale=0.4]{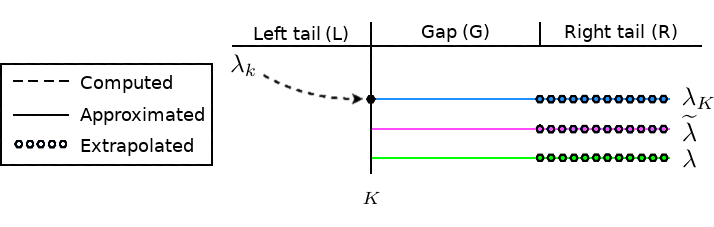}
	\caption{In terms of its eigenvalue spectrum, the covariance matrix can be partitioned as given by Equation (\ref{eq:eigendecomppart}): the left tail subspace (eigenpairs computed), the gap subspace (eigenvalues approximated, eigenvectors implicitly found by orthonormality) and the right tail subspace (eigenvalues extrapolated, eigenvectors implictly found by orthonormality).}
	\label{fig:gapanalysis}
\end{figure}

In terms of the Hessian and OPG estimators, the full eigendecomposition of the covariance matrix can be partitioned into three subspaces as shown in Figure \ref{fig:gapanalysis}
\begin{equation}
	\Sigma = \Sigma_{\text{L}} + \Sigma_{\text{G}} + \Sigma_{\text{R}}	= Q_{\text{L}}\Lambda_{\text{L}}^{-1}Q_{\text{L}}^T + Q_{\text{G}}\Lambda_{\text{G}}^{-1}Q_{\text{G}}^T + Q_{\text{R}}\Lambda_{\text{R}}^{-1}Q_{\text{R}}^T.\label{eq:eigendecomppart}
\end{equation}
This decomposition applies to both $\Sigma^\text{H}$ \eqref{eq:hess} and $\Sigma^\text{G}$ \eqref{eq:opg}, and thus we have omitted the superscripts in our notation. In practice, the two merely differs by which of the two matrices $H$ and $G$ the calculated eigenpairs come from. The subscript `G' denotes the gap subspace which is based on eigenvectors with eigenvalues $\lambda_{K + 1}$ to $\lambda_{P-K-1}$. Subscript `L' denotes the left tail subspace and is based on eigenvectors with eigenvalues $\lambda_1$ to $\lambda_{K}$. Finally, the subscript `R' denotes the right tail subspace which is based on eigenvectors with eigenvalues $\lambda_{P-K}$ to $\lambda_{P}$. Accordingly, we have that $Q_\text{L} \in \mathbb{R}^{P \times K}$, $\Lambda_{\text{L}} \in \mathbb{R}^{K \times K}$, $Q_\text{G} \in \mathbb{R}^{P \times (P - 2K)}$, $\Lambda_{\text{G}} \in \mathbb{R}^{(P - 2K)\times (P - 2K)}$, $Q_\text{R} \in \mathbb{R}^{P \times K}$ and $\Lambda_{\text{R}} \in \mathbb{R}^{P\times K}$.

If $\lambda_{K} \approx \lambda$ we can safely assume that all the eigenvalues in the gap subspace must be close to $\lambda$. In line with \cite{granziol} and the empirical evidence presented in Figure \ref{fig:eigenvaluespectrum}, we assume that all the eigenvalues in the right subspace are inherently noisy, and should not be used by the Hessian estimator. Consequently, we assume that also the eigenvalues in the right subspace are approximately equal to $\lambda$. Since the OPG estimator is always positive definite when $\lambda > 0$, the same assumption also holds true.

With reference to Figure \ref{fig:gapanalysis}, there are now two possible extreme conditions: a) when all the eigenvalues in the gap and right subspaces are set to $\lambda_{K}$ (blue), or b) when all the eigenvalues in the gap and right subspaces are set to $\lambda$ (green). By defining $\widetilde{\lambda}$ (purple) as the harmonic mean of $\lambda$ and $\lambda_K$, and $\epsilon_{\lambda}$ as the midpoint of their reciprocals,
\begin{equation}
	\widetilde{\lambda} = \left(\frac{\lambda^{-1} + \lambda_K^{-1}}{2}\right)^{-1} \quad \text{and} \quad
	\epsilon_{\lambda} = \frac{\lambda^{-1} - \lambda_K^{-1}}{2},
\end{equation}
it follows that $\widetilde{\lambda}^{-1} \pm \epsilon_{\lambda}$ will enclose the interval $[\lambda_K^{-1}, \lambda^{-1}]$. The covariance matrix can now be approximated by
\begin{align}
	\widetilde{\Sigma} &= \frac{1}{N}\left[Q_{\text{L}}\Lambda^{-1}_{\text{L}}Q_{\text{L}}^T + \widetilde{\lambda}^{-1} (Q_{\text{G}}Q_{\text{G}}^T + Q_{\text{R}}Q_{\text{R}}^T)\right],\label{eq:covapprox2}
\end{align}
with a worst-case approximation error $\Delta$ given by
\begin{align}
	\Delta &= \frac{\epsilon_{\lambda}}{N}\left[Q_{\text{G}}Q_{\text{G}}^T + Q_{\text{R}}Q_{\text{R}}^T\right],\label{eq:covapproxerror2}
\end{align}
such that $\Sigma$ is bounded by $\widetilde{\Sigma} \pm \Delta$. Since $Q$ is an orthonormal basis, we see that it is possible to express \eqref{eq:covapprox2} and \eqref{eq:covapproxerror2} without an explicit need to compute any of the eigenvectors relative to the gap nor right tail subspaces because
\begin{equation}
	Q_{\text{G}}Q_{\text{G}}^T + Q_{\text{R}}Q_{\text{R}}^T = I - Q_{\text{L}}Q_{\text{L}}^T.\label{eq:orthotrick}
\end{equation}
\noindent Inserting (\ref{eq:covapprox2}) into (\ref{eq:sigma1}) with use of \eqref{eq:orthotrick}, yields the final form of the approximation to the uncertainty associated with prediction of $x_0$
\begin{align}\label{eq:sigma3}
	\widetilde{\sigma}^2(x_0) &= \frac{1}{N}\text{diag}\left\{F\left[Q_{\text{L}}\Lambda^{-1}_{\text{L}}Q_{\text{L}}^T + \widetilde{\lambda}^{-1}(I - Q_{\text{L}}Q_{\text{L}}^T)\right]F^T\right\} \in \mathbb{R}^{T_L},
\end{align}
\noindent with a worst-case approximation error $\delta$ given by
\begin{equation}
	\delta = \frac{\epsilon_{\lambda}}{N}\text{diag}\left\{F\left(I - Q_{\text{L}}Q_{\text{L}}^T\right)F^T\right\}\in\mathbb{R}^{T_L},\label{eq:varerror}
\end{equation}
such that $\sigma^2(x_0)$ is bounded by $\widetilde{\sigma}^2(x_0) \pm \delta$.

In terms of standard deviations, the worst-case approximation error $\epsilon$ of $\widetilde{\sigma}(x_0)$ is given by
\begin{equation}
	\epsilon = \frac{1}{2}\left(\sqrt{\widetilde{\sigma}^2(x_0)+\delta} - \sqrt{\widetilde{\sigma}^2(x_0)-\delta} \right) \in\mathbb{R}^{T_L},\label{eq:error}
\end{equation}
such that $\sigma(x_0)$ is bounded by $\widetilde{\sigma}(x_0) \pm \epsilon$. Lastly, we define an `uncertainty score' (which we will use later to rank images) by summing the variances per class output (class variance), and then take the square root to get the total uncertainty in standard deviations
\begin{equation}
	\widetilde{\sigma}_\text{score}(x_0) = \sqrt{\sum_{m=1}^{T_L} {\widetilde{\sigma}^2_m}(x_0)} \in \mathbb{R},\label{eq:sigmascore}
\end{equation}
with the corresponding worst-case approximation error $\epsilon_\text{score}$ given by,
\begin{align}
	\epsilon_{\text{score}} = \frac{1}{2}\left(\sqrt{\sum_{m=1}^{T_L} {\widetilde{\sigma}^2_m(x_0)} + \delta_m} - \sqrt{\sum_{m=1}^{T_L} {\widetilde{\sigma}^2_m(x_0)} - \delta_m}\right) \in \mathbb{R},\label{eq:sigmascoreerror}
\end{align}
such that the true quantity is bounded by $\widetilde{\sigma}_\text{score}(x_0) \pm \epsilon_{\text{score}}$. We note that the worst-case approximation errors \eqref{eq:varerror}, \eqref{eq:error} and \eqref{eq:sigmascoreerror} are functions of $x_0$ but we have notationally dropped this from the equations to avoid cluttering. The approximation errors should be thought of as an uncertainty of the predictive uncertainty which accounts for the worst-case loss of not computing the gap subspace explicitly. Since the right tail subspace can be extrapolated when $H$ is not positive definite, the concept of an approximation error for the Hessian estimator must be used carefully.

At this point we make a few comments regarding Equation \eqref{eq:sigma3}. The first term on the right hand side, $Q_{\text{L}}\Lambda^{-1}_{\text{L}}Q_{\text{L}}^T$, corresponds to a low-rank approximation of the covariance matrix based on $K$ explicitly computed principal eigenpairs. However, when the second term, $\widetilde{\lambda}^{-1}(I - Q_{\text{L}}Q_{\text{L}}^T)$, is added -- the approximation becomes full-rank. When accounting for the left and right multiplication of the sensitivity matrix $F$, the per-class predictive uncertainties of $x_0$ can be interpreted as weighted sums of the squared sensitivities in the directions expressed by the eigenbasis $Q$ using the inverse eigenvalues as weights. Hence, for the low-rank approximation -- regardless of the sensitivity -- the contribution to the predictive uncertainty will be zero in directions $k > K$, whereas for the full-rank approximation -- the contribution can still be high. We will come back to this when we discuss out-of-distribution examples in Section \ref{sec:demo}.

\subsubsection{The Sandwich Estimator}
The approximation of the Sandwich estimator is defined by
\begin{equation}
	\widetilde{\Sigma} = \frac{1}{N}\widetilde{H}^{-1}\widetilde{G}\widetilde{H}^{-1}.\label{eq:sandwichapprox}
\end{equation}
We introduce two separate linearization constants for the approximation of the gap (and right tail) subspace of $G$ and $H^{-1}$ using the harmonic means
\begin{equation}
	\widetilde{\lambda}^{\text{H}} = \left(\frac{\lambda^{-1} + {\lambda_K^{\text{H}}}^{-1}}{2}\right)^{-1},\label{eq:lambdaGRH}
\end{equation}
\begin{equation}
	\widetilde{\lambda}^{\text{G}} = \left(\frac{\lambda^{-1} + {\lambda_K^{\text{G}}}^{-1}}{2}\right)^{-1}.\label{eq:lambdaGRG}
\end{equation}
\noindent The approximation of $H^{-1}$ is thus given by
\begin{equation}
	\widetilde{H}^{-1} = Q_{\text{L}}^{\text{H}}{\Lambda^{\text{H}}_\text{L}}^{-1}{Q_{\text{L}}^{\text{H}}}^T +
	{{}\widetilde{\lambda}^{\text{H}}}^{-1} (I - Q_{\text{L}}^{\text{H}}{Q_{\text{L}}^{\text{H}}}^T),\label{eq:approxH}
\end{equation}
and the approximation of $G$ given by
\begin{equation}
	\widetilde{G} = Q_{\text{L}}^{\text{G}}{\Lambda^{\text{G}}_\text{L}}{Q_{\text{L}}^{\text{G}}}^T + \widetilde{\lambda}^{\text{G}}(I - Q_{\text{L}}^{\text{G}}{Q_{\text{L}}^{\text{G}}}^T).\label{eq:approxG}
\end{equation}
The superscripts `H' and `G' are used to distinguish the eigenvectors and eigenvalues of $H$ and $G$ respectively.
By inserting \eqref{eq:approxH} and \eqref{eq:approxG} into \eqref{eq:sandwichapprox} and working out the product, we define the following eight matrices
\begin{align}
	\mathcal{S} &= Q_\text{L}^\text{H}{\Lambda_\text{L}^\text{H}}^{-1}{Q_\text{L}^\text{H}}^T Q_\text{L}^\text{G}\Lambda_\text{L}^\text{G}{Q_\text{L}^\text{G}}^T Q_\text{L}^\text{H}{\Lambda_\text{L}^\text{H}}^{-1}{Q_\text{L}^\text{H}}^T\\
	\mathcal{A} &= Q_\text{L}^\text{H}{\Lambda_\text{L}^\text{H}}^{-1}{Q_\text{L}^\text{H}}^T(I -  Q_\text{L}^\text{G}{Q_\text{L}^\text{G}}^T)Q_\text{L}^\text{H}{\Lambda_\text{L}^\text{H}}^{-1}{Q_\text{L}^\text{H}}^T\\
	\mathcal{N} &= (I - 
	Q_\text{L}^\text{H}{Q_\text{L}^\text{H}}^T )Q_\text{L}^\text{G}\Lambda_\text{L}^\text{G}{Q_\text{L}^\text{G}}^T Q_\text{L}^\text{H}{\Lambda_\text{L}^\text{H}}^{-1}{Q_\text{L}^\text{H}}^T\\
	\mathcal{D} &= (I - Q_\text{L}^\text{H}{Q_\text{L}^\text{H}}^T)(I - Q_\text{L}^\text{G}{Q_\text{L}^\text{G}}^T)Q_\text{L}^\text{H}{\Lambda_\text{L}^\text{H}}^{-1}{Q_\text{L}^\text{H}}^T\\
	\mathcal{W} &= Q_\text{L}^\text{H}{\Lambda_\text{L}^\text{H}}^{-1}{Q_\text{L}^\text{H}}^T Q_\text{L}^\text{G}\Lambda_\text{L}^\text{G}{Q_\text{L}^\text{G}}^T(I - Q_\text{L}^\text{H}{Q_\text{L}^\text{H}}^T) = \mathcal{N}^T\\
	\mathcal{I} &= Q_\text{L}^\text{H}{\Lambda_\text{L}^\text{H}}^{-1}{Q_\text{L}^\text{H}}^T(I - Q_\text{L}^\text{H}{Q_\text{L}^\text{H}}^T)(I - Q_\text{L}^\text{G}{Q_\text{L}^\text{G}}^T) = \mathcal{D}^T\\
	\mathcal{C} &= (I - Q_\text{L}^\text{H}{Q_\text{L}^\text{H}}^T)Q_\text{L}^\text{G}\Lambda_\text{L}^\text{G}{Q_\text{L}^\text{G}}^T(I - Q_\text{L}^\text{H}{Q_\text{L}^\text{H}}^T)\\
	\mathcal{H} &= (I - Q_\text{L}^\text{H}{Q_\text{L}^\text{H}}^T)(I - Q_\text{L}^\text{G}{Q_\text{L}^\text{G}}^T)(I - Q_\text{L}^\text{H}{Q_\text{L}^\text{H}}^T)
\end{align}
The uncertainty associated with prediction of $x_0$ can now be written
\begin{align}
	\widetilde{\sigma}^2(x_0) = \frac{1}{N}\text{diag}\big\{F\big[\mathcal{S} &+ \widetilde{\lambda}^\text{G}\mathcal{A}\nonumber\\ &+ {{}\widetilde{\lambda}^{\text{H}}}^{-1}(\mathcal{N} + \mathcal{N}^T)\nonumber\\ &+ \widetilde{\lambda}^\text{G}{{}\widetilde{\lambda}^{\text{H}}}^{-1}(\mathcal{D} + \mathcal{D}^T)\nonumber\\&+ {{}\widetilde{\lambda}^{\text{H}}}^{-2}\mathcal{C}\nonumber\\&+ \widetilde{\lambda}^\text{G}{{}\widetilde{\lambda}^{\text{H}}}^{-2}\mathcal{H}\big]F^T\big\} \in \mathbb{R}^{T_L},\label{eq:sigma4_5}
\end{align}
with the worst-case approximation error given by
\begin{align}
	\delta = \frac{1}{2N}\text{diag}\big\{&F\big[ (\lambda_{K}^{\text{G}} - \lambda)\mathcal{A}\nonumber\\ &+ (\lambda^{-1} - {\lambda_{K}^{\text{H}}}^{-1})(\mathcal{N} + \mathcal{N}^T)\nonumber\\ &+ (\lambda_{K}^{\text{G}}\lambda^{-1} - {\lambda_{K}^{\text{H}}}^{-1}\lambda)(\mathcal{D} + \mathcal{D}^T)\nonumber\\ &+ (\lambda^{-2} - {\lambda_{K}^{\text{H}}}^{-2})\mathcal{C}\nonumber\\ &+ (\lambda^{-2}\lambda_{K}^{\text{G}} - {{\lambda_{K}^{\text{H}}}}^{-2}\lambda)\mathcal{H} \big]F^T \big\}\in \mathbb{R}^{T_L},\label{eq:varerror2}
\end{align}
such that $\sigma^2(x_0)$ is bounded by $\widetilde{\sigma}^2(x_0) \pm \delta$.
\noindent In terms of standard deviations, the approximation error is readily found by inserting \eqref{eq:sigma4_5} and \eqref{eq:varerror2} into \eqref{eq:error}.

\subsection{On the Relation Between the Effective Number of Parameters and $K$}
In \cite{mackay}, the so-called effective number of parameters is defined in terms of the eigenvalues of the Hessian matrix. It is noted that directions in parameter space for which the eigenvalues are close to $\lambda$ do not contribute to the number of good parameter measurements. Therefore, the effective number of parameters is a measure of the number of parameters which are well determined by the training data. In other words, when we select $K$ so that $\lambda_K \approx \lambda$, we loosely cover the data dependent part of the Hessian matrix (first term of right hand side of (\ref{eq:costfunction})) and can therefore expect that $K$ will be a crude estimate of the number of effective parameters.

As seen by Equations \eqref{eq:varerror} and \eqref{eq:varerror2}, the approximation error will be zero when the smallest eigenvalue $\lambda_K$ in the left tail subspace (of $H$ and $G$) is exactly equal to the $L_2$-regularization rate $\lambda$.

\section{Demonstration and Proof of Concept}\label{sec:demo}
In the following Section we explore and demonstrate the approximate predictive epistemic uncertainty estimate governed by \eqref{eq:sigma1} for the two LeNet-based neural network classifiers that were introduced in Section \ref{sec:classifier}. We establish by the use of regressions that the three estimators \eqref{eq:hess}-\eqref{eq:sandwich} yield close to perfectly correlated predictive epistemic uncertainty estimates for both of the classifiers.

\subsection{The Distribution of Approximate Predictive Epistemic Uncertainty}
\begin{figure}[h]
	\centering
	\begin{subfigure}{0.33\textwidth}
		\includegraphics[scale=0.19]{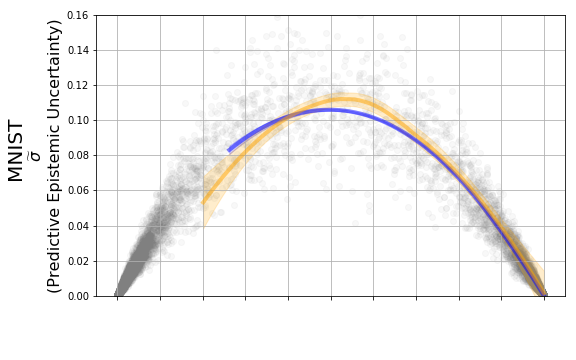}
		\label{fig:mnistuncertaintydistributiontestH}
	\end{subfigure}%
	\begin{subfigure}{0.33\textwidth}
		\includegraphics[scale=0.19]{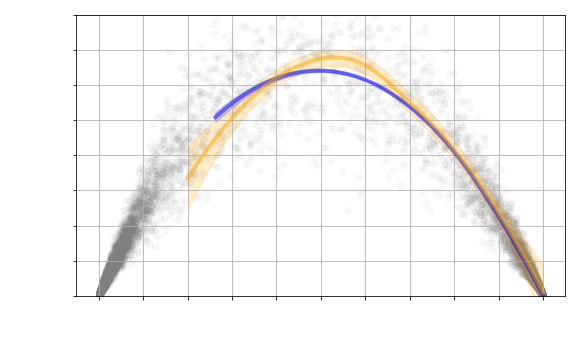}
		\label{fig:mnistuncertaintydistributiontestG}
	\end{subfigure}%
	\begin{subfigure}{0.33\textwidth}
		\includegraphics[scale=0.19]{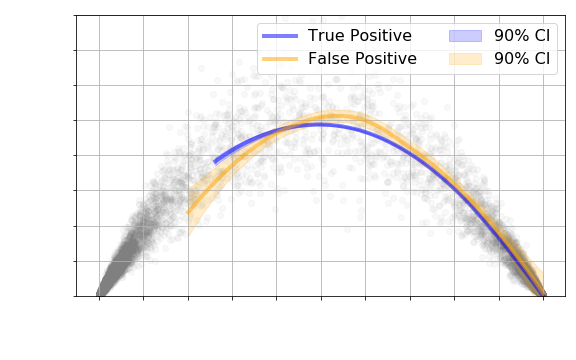}
		\label{fig:mnistuncertaintydistributiontestS}
	\end{subfigure}
	\begin{subfigure}{0.33\textwidth}
		\includegraphics[scale=0.19]{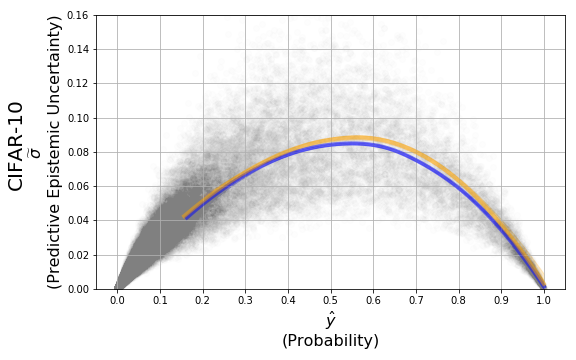}
		\caption{Hessian}
		\label{fig:cifar-10uncertaintydistributiontestH}
	\end{subfigure}%
	\begin{subfigure}{0.33\textwidth}
		\includegraphics[scale=0.19]{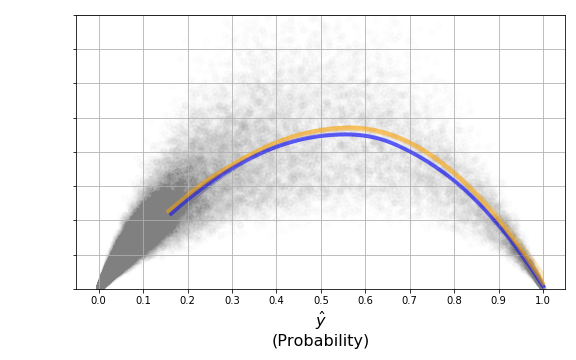}
		\caption{OPG}
		\label{fig:cifar-10uncertaintydistributiontestG}
	\end{subfigure}%
	\begin{subfigure}{0.33\textwidth}
		\includegraphics[scale=0.19]{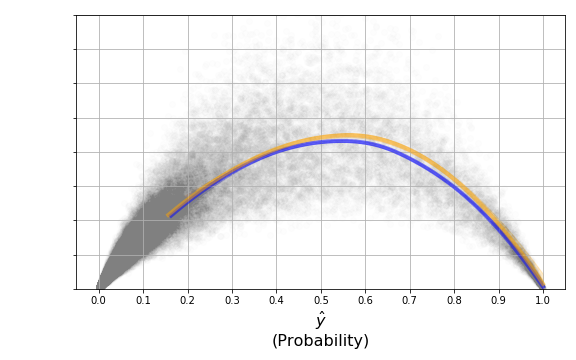}
		\caption{Sandwich}
		\label{fig:cifar-10uncertaintydistributiontestS}
	\end{subfigure}
	\caption{Nonparametrically smoothed versions of the predictive epistemic uncertainty \eqref{eq:sigma1} for the true positives (blue) and false positives (orange) in the MNIST (upper row, $K=1500$) and CIFAR-10 (lower row, $K=2500$) test sets as functions of class probability for each of the three estimators. The shaded gray bullets ($N \times T_L$ such bullets) represent the raw predictive uncertainty for all $T_L$ classes against probability.}
	\label{fig:uncertaintydistributiontest}
\end{figure}

Figure \ref{fig:uncertaintydistributiontest} shows nonparametrically smoothed versions of the predictive epistemic uncertainty for the three proposed estimators against class probability for all the images in the MNIST and CIFAR-10 test sets. Clearly, the three estimators yield close to identical results. Further, we observe that the average predictive epistemic uncertainty associated with false positives (yellow line) is higher than for true positives (blue line). The banana-shaped appearance of these plots suggests that there is a negative quadratic relationship between probability and uncertainty. The explanation for this is attributed to the softmax activation function whose gradient (e.g. sensitivity $F$) will always be weighted by a quantity which is negative quadratic in probability (e.g. $\hat{y}(1-\hat{y})$).

\begin{figure}[h]
	\centering
	\begin{subfigure}{.5\textwidth}
		\centering
		\includegraphics[scale=0.25]{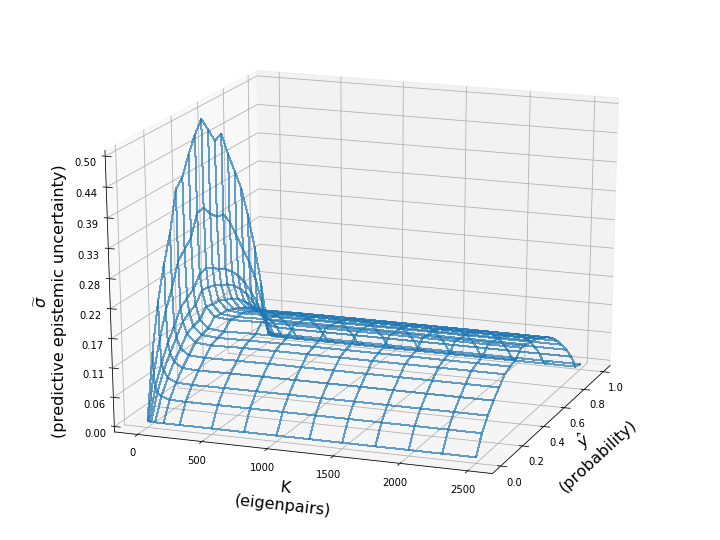}
		\label{fig:mnisttest3dsigma}
	\end{subfigure}%
	\begin{subfigure}{.5\textwidth}
		\centering
		\includegraphics[scale=0.25]{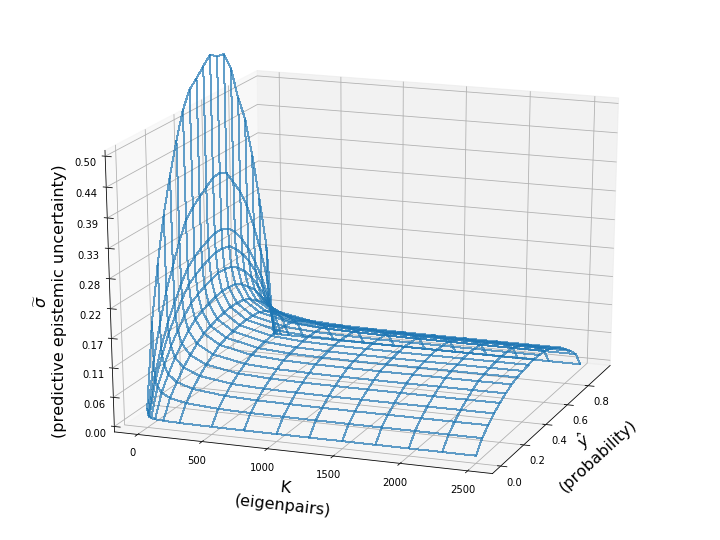}
		\label{fig:mnisttest3dsigma}
	\end{subfigure}
	\begin{subfigure}{.5\textwidth}
		\centering
		\includegraphics[scale=0.25]{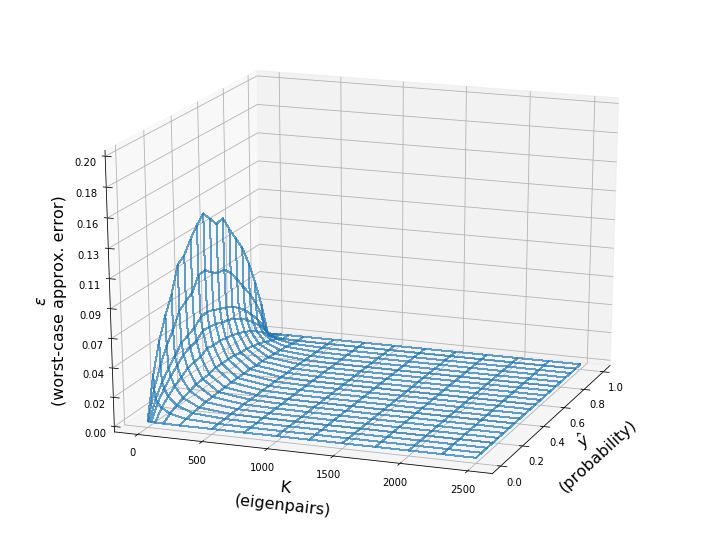}
		\caption{MNIST}
		\label{fig:mnisttest3depsilon}
	\end{subfigure}%
	\begin{subfigure}{.5\textwidth}
		\centering
		\includegraphics[scale=0.25]{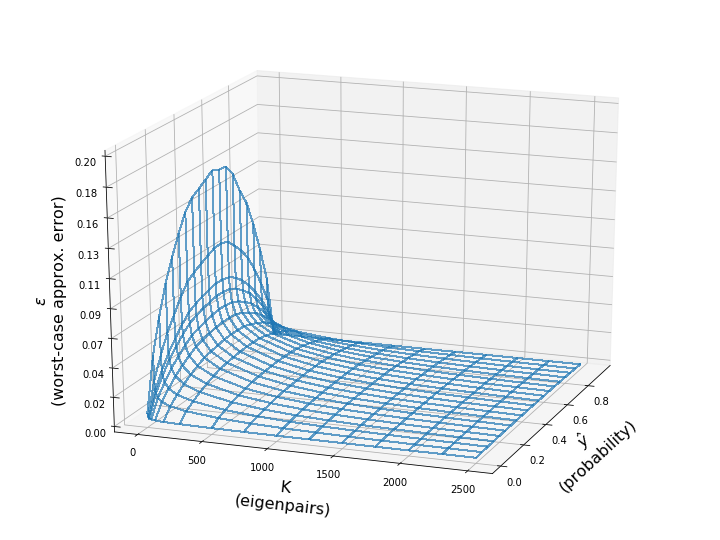}
		\caption{CIFAR-10}
		\label{fig:mnisttest3depsilon}
	\end{subfigure}
	\caption{Nonparametrically smoothed versions of the predictive epistemic uncertainty (upper row) and the  approximation error (lower row) in the MNIST and CIFAR-10 test sets as functions of the number of computed eigenpairs $K$ and class probability using the OPG estimator.}
	\label{fig:test3d}
\end{figure}

The evolution of the nonparametrically smoothed uncertainty levels and approximation errors for the OPG estimator as functions of the number of computed eigenpairs $K$ and class probability is displayed in Figure \ref{fig:test3d}. As expected, for a growing $K$, the approximation errors diminish and the uncertainty stabilizes. Although we do not display similar plots for the other two estimators, we note that for MNIST, the approximation errors are smallest for the OPG estimator, followed by the Hessian estimator and the Sandwich estimator. The larger the difference between $\lambda$ and the smallest eigenvalue $\lambda_K$, the higher the average approximation error. As seen by the eigenvalue spectra in Figure \ref{fig:eigenvaluespectrum}, the drop-off rate towards $\lambda$ is faster for $G$, thus explaining why the OPG estimator leads to the lowest approximation errors on MNIST. We note that since the Sandwich estimator is dependent on both the approximation of $H$ and $G$, its approximation errors are not unexpectedly the highest. Furthermore, the fall-off rate towards $\lambda$ in the eigenvalue spectrum for CIFAR-10 is slightly lower than for MNIST. This means that the CIFAR-10 classifier has a greater number of effective parameters -- and thus requires a higher $K$ to achieve acceptable approximation error levels. This fact is evident by Figure \ref{fig:test3d}, where we see that the OPG approximation errors for CIFAR-10 are dropping off to zero slower than for MNIST.

For all three estimators, it is evident by Figure \ref{fig:test3d} that most of the contribution to the predictive epistemic uncertainty comes from the left subspace corresponding to the largest eigenvalues of $H$ and $G$. This observation can be counter-intuitive since it is the directions with the smallest eigenvalues that will be the largest contributors to the variance when accounting for the inversions in \eqref{eq:hess}, \eqref{eq:opg} or \eqref{eq:sandwich}.

The explanation for this phenomenon is attributed to the sensitivity $F$ \eqref{eq:jacobian}. We observe that the training and test set sensitivity drops to zero in directions $k$ for which $\lambda_k \approx \lambda$ and is thus canceling with the reciprocals of the smallest eigenvalues in the linear combinations formed by \eqref{eq:sigma3} or \eqref{eq:sigma4_5}. Nevertheless, as the sensitivity for data not belonging to the same distribution as the training can still be high in these directions, the corresponding predictive epistemic uncertainty can still receive significant contributions from directions $k > K$. This emphasizes the importance of making the estimators full-rank using the orthonormal basis technique presented in Section \ref{sec:closingthegap}. We add that due to the full-rank property, the number $K$ should be thought of as the number of explicitly computed eigenpairs rather than the number of utilized eigenpairs -- as the latter will effectively be equal to $P$. 

\begin{figure}[h]
	\centering
	\begin{subfigure}{.7\textwidth}
		\centering
		\includegraphics[scale=0.38]{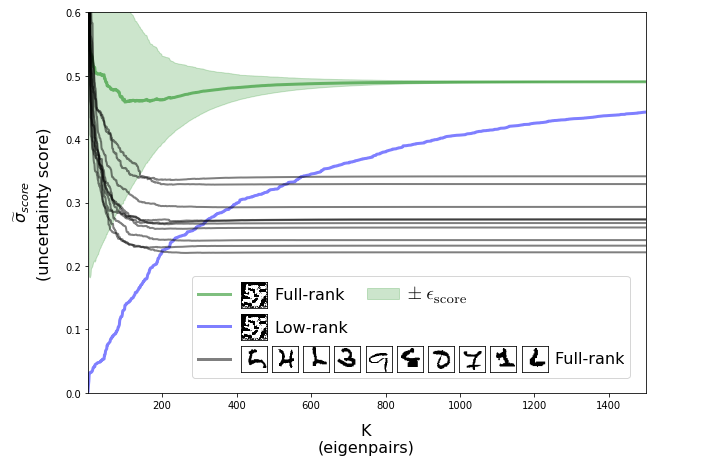}
		\caption{MNIST Uncertainty Scores}
		\label{fig:oodevo}
	\end{subfigure}%
	\begin{subfigure}{.3\textwidth}
		\centering
		\includegraphics[scale=0.33]{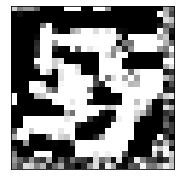}
		\caption{\\MNIST\\OoD Example}
		\label{fig:oodimage}
	\end{subfigure}
	\caption{The uncertainty score (a) as a function of $K$ for the MNIST OoD example in (b) using the full-rank OPG approximation (green curve) vs. its low-rank counterpart (blue curve) from Equations \eqref{eq:sigma3} and \eqref{eq:sigmascore}. The green interval corresponds to the approximation error. The reference images (black curves) are computing using the full-rank approximation, and corresponds to the ten images in the training set with the highest uncertainty scores sorted in descending order.}
	\label{fig:ood}
\end{figure}

To illustrate the concept of a low vs. full-rank approximation, Figure \ref{fig:oodevo} displays the uncertainty scores as functions of $K$ for the low and full-rank version of the OPG estimator applied to the out-of-distribution (OoD) example shown in Figure \ref{fig:oodimage}. For reference, we also plot the uncertainty scores for the ten images in the training set with the highest uncertainty scores sorted in descending order. Comparing the green curve with the blue curve shows that the OoD example has a sensitivity spectrum stretching out far beyond $K=1500$ because the low-rank version (blue) has not yet reached the stable level achieved by the full-rank approximation (green) at this $K$. That the full-rank approximation quickly stabilizes already at around $K=600$, can be explained by that it receives contribution from the full spectrum even though only $K$ principal eigenpairs are computed explicitly at each stage. The reference images (black curves) are computing using the full-rank approximation, and are all lower ranked than the OoD example.
A detailed comparison of the three estimators is shown in Table \ref{tab:corr}. By regressing their outcomes against each other, we clearly see that the relative estimated uncertainty levels are near to perfectly correlated since the squared correlations coefficients are close to $1$. As seen by the slopes $\beta$, only the absolute levels of the estimated uncertainty differ, and since the intercepts $\alpha$ are zero, there are no offsets. 

\begin{table}[h]
	\scalebox{0.7}{
		\begin{tabular}{lcccccccccc}
			& \multicolumn{1}{l}{}                                            & \multicolumn{3}{c}{\shortstack{\\\textbf{Hessian vs. OPG} \\ \boldmath$\widetilde{\sigma}^{\text{G}}(x_n) = \alpha + \beta\widetilde{\sigma}^{\text{H}}(x_n)$\unboldmath}} &\multicolumn{3}{c}{\shortstack{\\\textbf{Hessian vs. Sandwich} \\ \boldmath$\widetilde{\sigma}^{\text{S}}(x_n) = \alpha + \beta\widetilde{\sigma}^{\text{H}}(x_n)$\unboldmath}}           & \multicolumn{3}{c}{\shortstack{\\\textbf{OPG vs. Sandwich} \\ \boldmath$\widetilde{\sigma}^{\text{S}}(x_n) = \alpha + \beta\widetilde{\sigma}^{\text{G}}(x_n)$\unboldmath}} \\ \cline{3-11} 
			& \multicolumn{1}{l}{}                                            & $\mathbf{R^2}$ & $\pmb{\alpha}$ & $\pmb{\beta}$ & $\mathbf{R^2}$ & $\pmb{\alpha}$ & $\pmb{\beta}$ &  $\mathbf{R^2}$ & $\pmb{\alpha}$ & $\pmb{\beta}$            \\ \hline
			\multicolumn{1}{c|}{\multirow{2}{*}{\textbf{MNIST}}}    & \textbf{\shortstack{\\Training\\ Set}} &  0.997                              & 0.000                              & 1.206    &  0.998          & 0.000          & 0.923         & 0.990          & 0.000          & 0.761 \\
			\multicolumn{1}{c|}{}                                   & \textbf{\shortstack{\\Test\\ Set}}     &  0.998                              & 0.000                              & 1.219                               & 0.999          & 0.000          & 0.915          & 0.995          & 0.000          & 0.748\\ \hline
			\multicolumn{1}{l|}{\multirow{2}{*}{\textbf{CIFAR-10}}} & \textbf{\shortstack{\\Training\\ Set}} &  0.999            & 0.000              &  1.062            &    0.999              &      0.000          & 1.017               &    0.997            &  0.000             &  0.956 \\
			\multicolumn{1}{l|}{}                                   & \textbf{\shortstack{\\Test\\ Set}}     &   1.000            & 0.000              & 1.066             &    1.000              &      0.000          &  1.014               & 0.998               & 0.000              &  0.950   \\ \hline
		\end{tabular}
		}
		\caption{Regression comparison of $\widetilde{\sigma}^{\text{H}}$, $\widetilde{\sigma}^{\text{G}}$ and $\widetilde{\sigma}^{\text{S}}$ across all the images in the MNIST and CIFAR-10 training and test sets. The respective superscripts $H$, $G$ and $S$ denote Hessian, OPG and Sandwich. The regression intercept, slope and squared correlation coefficient is denoted by $\alpha$, $\beta$ and $R^2$, respectively.}
		\label{tab:corr}
\end{table}

\subsection{Ranking Images Based on the `Uncertainty Score'}\label{sec:ranking}
We propose to validate our results by studying the MNIST and CIFAR-10 images associated with the maximum and minimum amount of total predictive epistemic uncertainty as defined in (\ref{eq:sigmascore}) using the Hessian estimator. Unsurprisingly, since the squared correlation coefficients in Table \ref{tab:corr} are close to 1, the OPG and Sandwich estimators yield almost identical results and are not shown. 

The idea is based on the following reasoning: if a neural network classifies an image with low predictive epistemic `uncertainty score', the image should be easy to classify also for a human. Conversely, if the neural network classifies an image with a high predictive epistemic `uncertainty score', the image should be hard to classify for a human. Effectively, the predictive epistemic `uncertainty score' ranks images according to the degree of `doubt' expressed by the neural network -- and by the figures we find striking evidence that this corresponds well with human judgment.

\begin{figure*}[h]
	\centering
	\begin{subfigure}{0.25\textwidth}
		\centering
		\boxed{
		\includegraphics[scale=0.15]{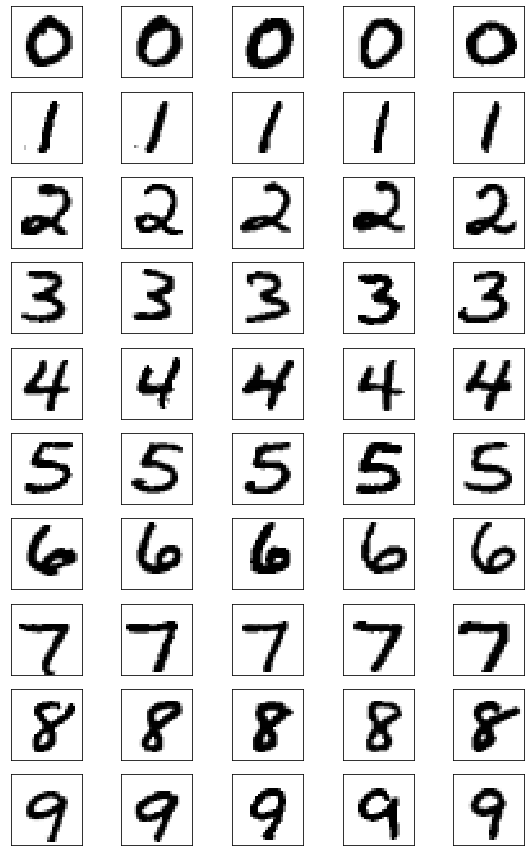}
		}
		\label{fig:mnistgreatestuncertaintyimagestrain}
	\end{subfigure}%
	\begin{subfigure}{0.25\textwidth}
	\centering
	\boxed{
	\includegraphics[scale=0.15]{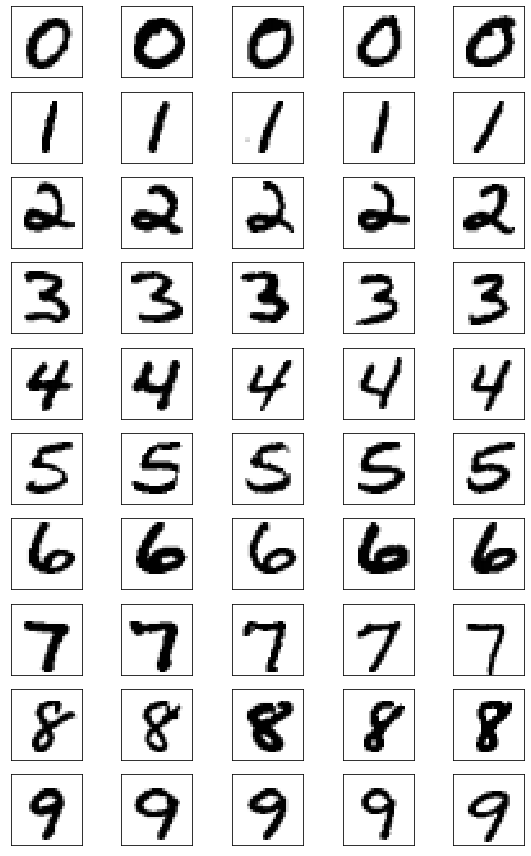}
	}
	\label{fig:mnistgreatestuncertaintyimagestrain}
\end{subfigure}%
	\begin{subfigure}{0.25\textwidth}
		\centering
		\boxed{
		\includegraphics[scale=0.15]{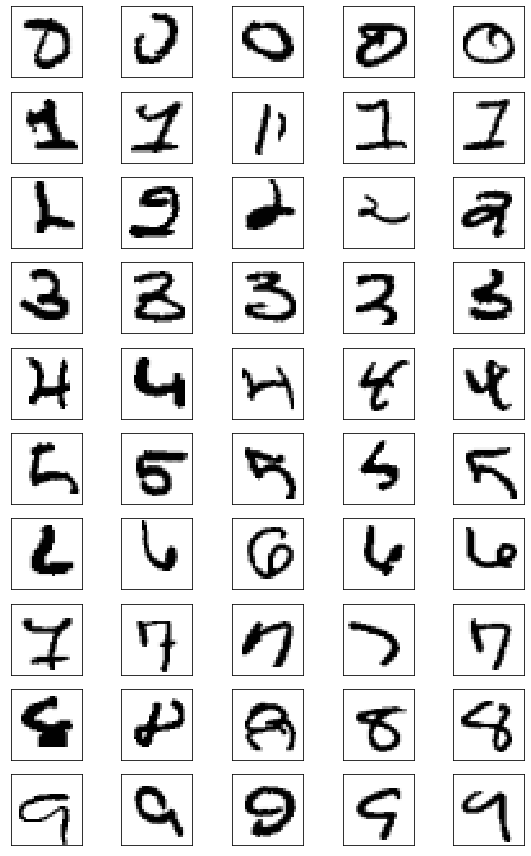}
		}
		\label{fig:mnistlowestuncertaintyimagestrain}
	\end{subfigure}%
	\begin{subfigure}{0.25\textwidth}
		\centering
		\boxed{
		\includegraphics[scale=0.15]{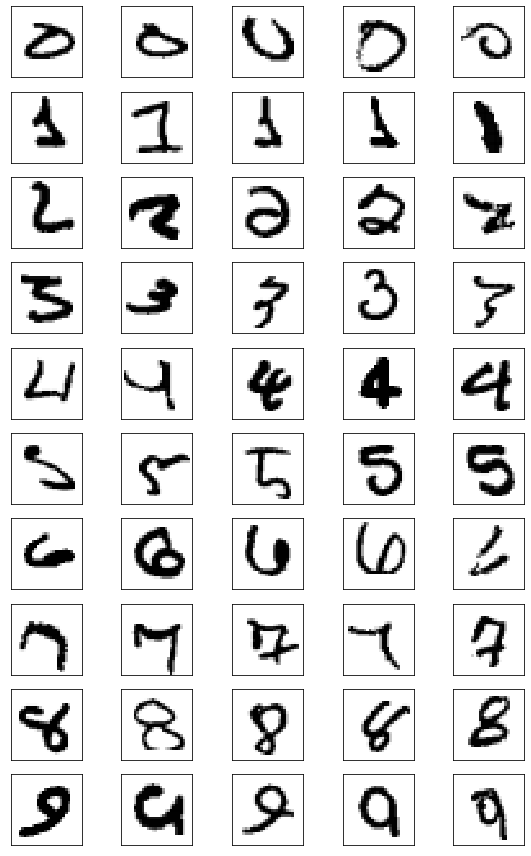}
		}
		\label{fig:mnistlowestuncertaintyimagestrain}
	\end{subfigure}
	\begin{subfigure}{0.25\textwidth}
	\centering
	\boxed{
		\includegraphics[scale=0.15]{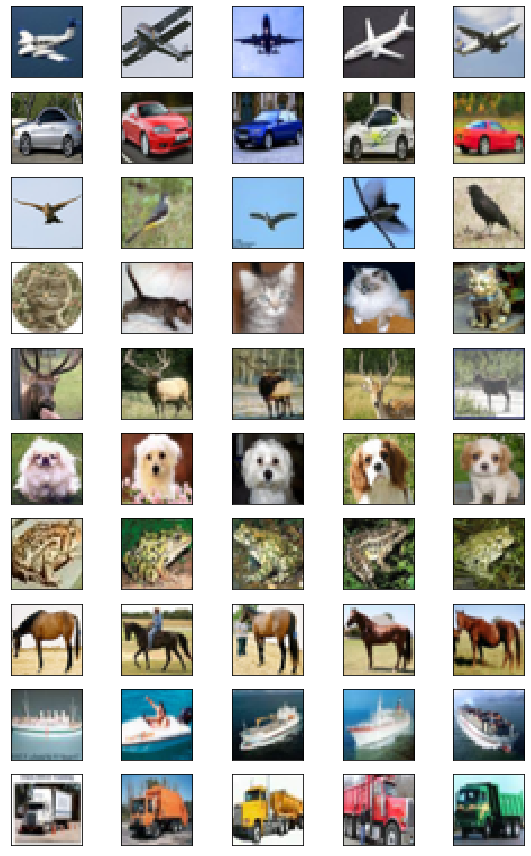}
	}
	\caption{}
	\label{fig:cifar-10greatestuncertaintyimagestrain}
\end{subfigure}%
\begin{subfigure}{0.25\textwidth}
	\centering
	\boxed{
		\includegraphics[scale=0.15]{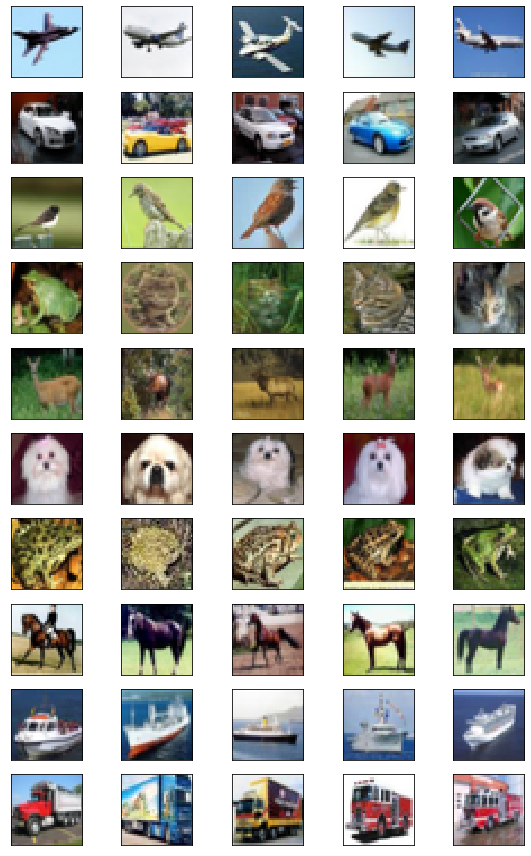}
	}
	\caption{}
	\label{fig:cifar-10greatestuncertaintyimagestrain}
\end{subfigure}%
\begin{subfigure}{0.25\textwidth}
	\centering
	\boxed{
		\includegraphics[scale=0.15]{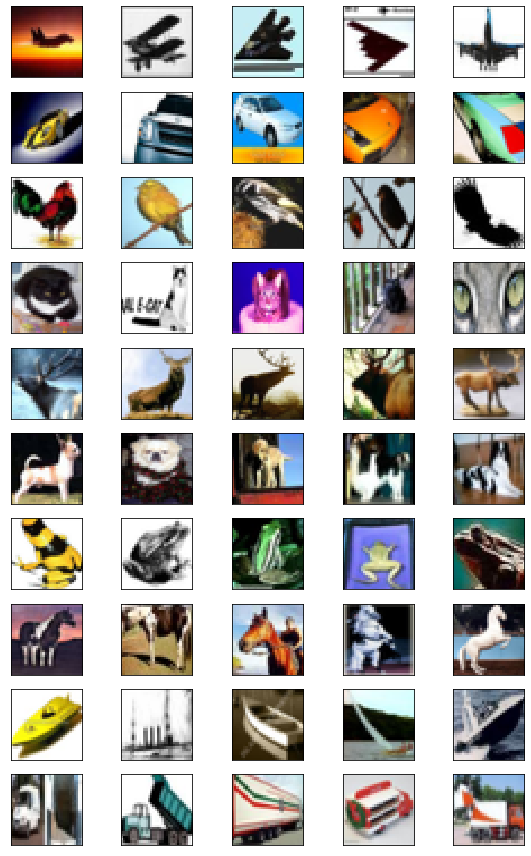}
	}
	\caption{}
	\label{fig:cifar-10lowestuncertaintyimagestrain}
\end{subfigure}%
\begin{subfigure}{0.25\textwidth}
	\centering
	\boxed{
		\includegraphics[scale=0.15]{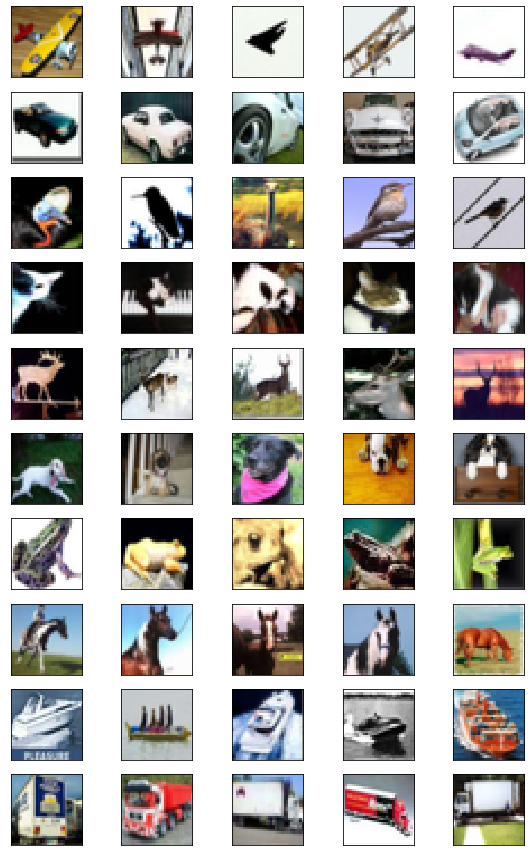}
	}
	\caption{}
	\label{fig:cifar-10lowestuncertaintyimagestrain}
\end{subfigure}%
	\caption{The MNIST and CIFAR-10 images ranked by the predictive epistemic `uncertainty score' per class: (a) lowest 5 in the training set, (b) lowest 5 in the test set, (c) highest 5 in the training set and (d) highest 5 in the test set.}
\end{figure*}

\section{Summary, Concluding Remarks and Further Work}\label{sec:summary}
We have presented a computationally tractable framework for traditional Fisher information based uncertainty quantification in deep learning classification. To this end, we have introduced full-rank, positive definite covariance estimators using approximate eigendecompositions in terms of either the Hessian, the OPG approximation or the so-called Sandwich estimator. Further, we have proposed to utilize the Lanczos algorithm in combination with Pearlmutter's technique to compute the needed eigenpairs of the Hessian, and to compute mini-batches of the Jacobian matrix using efficient per-example gradients in combination with incremental singular value decompositions for the OPG approximation. As the computational complexity of these methods scale linearly with the number of model parameters, they are therefore suited for deep learning.

We have shown that the three estimators yield close to identical prediction uncertainty estimates when applied on two different LeNet-based neural network classifiers. We have seen that only the top $K << P$ Fisher information matrix eigenpairs contribute significantly to the predictive uncertainty for data in the same distribution as the training set. As this does not necessarily hold true for OoD examples, we have shown that thanks to the full-rank property of the proposed estimators, also these will converge quickly under the same framework.

We have also seen that when images are ranked according to their relative level of predictive epistemic uncertainty, the ordering corresponds well with human judgment: corner cases tend to be highly ranked, and we clearly see why data augmentation is beneficial -- since the top ranked images often are prone to unusual perspectives and/or rare colors. Generally, we conjecture that classifiers can benefit from operating in the joint probability-uncertainty domain. As a corroborative example we have empirically shown that false positives appears to have an average higher prediction uncertainty than true positives.

Looking forward, we point at several specific areas of research which could be investigated. The first candidate is to establish how the Fisher information eigenspectrum of very large networks and datasets behave. If the contraction of the spectrum towards $\lambda$ continues to be fast with growing network and dataset sizes, the methodology presented can be tractable even for the most complex models. However, if the largest affordable $K$ yields a $\lambda_K$ far from $\lambda$, it can render the methodology intractable in terms of that the approximation errors will be too large. This points to understanding what causes the contraction phase in the first place, and hence uncovering the factors that drive it. Secondly, we leave the discussion regarding which of the three estimators (or other combinations) one should use -- and when -- as an opportunity for future research. Thirdly, as this work has been focused on the classification task, a natural extension is to see how the framework behaves under deep learning regression \cite{khosravi}. Fourthly, we point at a fundamental issue with the Delta method itself. The Delta method is inevitably based on the local curvature around the parameter estimate $\hat{\omega}$, hence incorporating no means about the uncertainty of the parameter estimate outside this local region. What is lost, and how much, by disregarding the broader perspective of the solution space -- a space potentially within reach for sampling methods. Finally, we hope that this contribution and the released software can pave the way for not only uncertainty focused research, but for a broader range of Hessian based research topics in the deep learning domain.

\section{Acknowledgements}
Parts of this work have been done in the context of CEDAS (Center for Data Science, University of Bergen, Norway). The lead author would also like to thank Dr. Berent \AA. S. Lunde and \O yvind Lunde R\o rtveit for their helpful advice on various technical issues examined in this paper.

\newpage
\section{Appendix}\label{sec:appendix}
The cost function $C(\omega)$ can be interpreted as the negative log posterior,
\begin{equation}
	C(\omega) = -\log p(\mathcal{D}|\omega)p(\omega),
\end{equation}
for the parameter $\omega$ and some training data $\mathcal{D}$, where $p(\mathcal{D}|\omega)$ is the likelihood and
$p(\omega)$ the prior. Under $L_2$-regularization with rate $\lambda/2$, the prior takes the
form of a multivariate normal distribution with zero mean and covariance
$(\lambda/2)^{-1}I$
\begin{equation}
	\omega \sim \mathcal{N}\big(0, (\lambda/2)^{-1}I\big).
\end{equation}
It follows that
\begin{equation}
	H_{C(\omega)} = -H_{\log p(\mathcal{D}|\omega)p(\omega)}
	=-H_{\log p(\mathcal{D}|\omega)}+\lambda I,
\end{equation}
where we have used that $H_{\log p(\omega)}=-\lambda I$. Taking expectation with respect to $p(\mathcal{D}|\omega)$,
and drawing on the well known result for the expected Fisher information matrix \cite{casella}:
\begin{equation}
	\underset{{p(\mathcal{D}|\omega)}}{\mathbb{E}} \left[H_{\log p(\mathcal{D}|\omega)}\right]
	= - \underset{{p(\mathcal{D}|\omega)}}{\mathbb{E}}\left[ \nabla \log
	p(\mathcal{D}|\omega) \nabla \log p(\mathcal{D}|\omega)^T\right],
\end{equation}
it follows that
\begin{equation}
	\underset{{p(\mathcal{D}|\omega)}}{\mathbb{E}}\left[H_{C(\omega)}\right]
	=
	\underset{{p(\mathcal{D}|\omega)}}{\mathbb{E}}\left[G\right]
	+\lambda I\QED
\end{equation}

\bibliographystyle{abbrv}
\bibliography{references}

\begin{thebibliography}{10}

\bibitem{loquercio}
M.~S. A.~Loquercio and D.~Scaramuzza.
\newblock {A General Framework for Uncertainty Estimation in Deep Learning}.
\newblock \url{https://arxiv.org/pdf/1907.06890}, arXiv:1907.06890v4 [cs.CV],
  2020.

\bibitem{alain}
G.~Alain, N.~L. Roux, and P.-A. Manzagol.
\newblock {Negative eigenvalues of the Hessian in deep neural networks}.
\newblock \url{https://arxiv.org/abs/1902.02366}, arXiv:1902.02366v1 [cs.LG],
  2019.

\bibitem{bottou}
L.~Bottou, F.~E. Curtis, and J.~Nocedal.
\newblock Optimization methods for large-scale machine learning.
\newblock SIAM Rev., vol. 60, no. 2, pp. 223-311, 2018.

\bibitem{cardot}
H.~Cardot and D.~Degras.
\newblock {Online Principal Component Analysis in High Dimension: Which
  Algorithm to Choose?}
\newblock \url{https://arxiv.org/abs/1511.03688} arXiv:1511.03688 [stat.ML],
  2015.

\bibitem{freedman}
D.~A. Freedman.
\newblock {On the So-Called "Huber Sandwich Estimator" and "Robust Standard
  Errors"}.
\newblock \url{https://www.jstor.org/stable/27643806} The American
  Statistician, Vol. 60, No. 4 (Nov., 2006), pp. 299-302, 2006.

\bibitem{gal}
Y.~Gal and Z.~Ghahramani.
\newblock {Dropout as a Approximation: Representing Model Uncertainty in Deep
  Learning}.
\newblock \url{https://arxiv.org/pdf/1506.02142}, arXiv:1506.02142v6 [stat.ML],
  2016.

\bibitem{ghorbani}
B.~Ghorbani, S.~Krishnan, and Y.~Xiao.
\newblock {An Investigation into Neural Net Optimization via Hessian Eigenvalue
  Density}.
\newblock \url{https://arxiv.org/abs/1901.10159}, arXiv:1901.10159v1 [cs.LG],
  2019.

\bibitem{goodfellowetal}
I.~Goodfellow, Y.~Bengio, and A.~Courville.
\newblock {Deep Learning}.
\newblock \url{http://www.deeplearningbook.org}, MIT Press, 2016.

\bibitem{granziol}
D.~Granziol, T.~Garipov, S.~Zohren, D.~Vetrov, S.~Roberts, and A.~G. Wilson.
\newblock {The Deep Learning Limit}: are negative neural network eigenvalues
  just noise?
\newblock Presented at the ICML 2019 Workshop on Theoretical Physics for Deep
  Learning, 2019.

\bibitem{grosse}
R.~Grosse.
\newblock {Lecture 2: Taylor Approximations}.
\newblock
  \url{https://www.cs.toronto.edu/~rgrosse/courses/csc2541_2021/readings/L02_Taylor_approximations.pdf},
  2020.

\bibitem{hoef}
J.~M.~V. Hoef.
\newblock {Who Invented the Delta Method?}
\newblock
  \url{https://www.researchgate.net/publication/254329376_Who_Invented_the_Delta_Method},
  The American Statistician, 66:2, 124-127, 2012.

\bibitem{hullermeier}
E.~H\"ullermeier and W.~Waegeman.
\newblock {Aleatoric and Epistemic Uncertainty in Machine Learning: An
  Introduction to Concepts and Methods}.
\newblock \url{https://arxiv.org/abs/1910.09457}, arXiv:1910.09457v2 [cs.LG],
  2020.

\bibitem{gal2}
A.~Kendall and Y.~Gal.
\newblock {What Uncertainties Do We Need in Bayesian Deep Learning for Computer
  Vision?}
\newblock \url{https://arxiv.org/pdf/1703.04977}, arXiv:1703.04977v2 [cs.CV],
  2017.

\bibitem{khosravi}
A.~Khosravi and D.~Creighton.
\newblock {A Comprehensive Review of Neural Network-based Prediction Intervals
  and New Advances}.
\newblock
  \url{https://www.researchgate.net/publication/51534965_Comprehensive_Review_of_Neural_Network-Based_Prediction_Intervals_and_New_Advances},
  IEEE Transactions On Neural Networks, Vol. 22, No. 9, 2011.

\bibitem{kingma}
D.~P. Kingma and J.~L. Ba.
\newblock {Adam: A method for stochastic optimization}.
\newblock In Proc. 3rd Int. Conf. Learn. Representations, 2014.

\bibitem{lecun2}
Y.~LeCun, P.~Y. Simard, and B.~Pearlmutter.
\newblock {Automatic Learning Rate Maximization by On-Line Estimation of the
  Hessian's Eigenvectors}.
\newblock
  \url{http://yann.lecun.com/exdb/publis/pdf/lecun-simard-pearlmutter-93.pdf},
  Advances in Neural Information Processing Systems (NIPS 1992), 1993.

\bibitem{casella}
E.~L. Lehmann and G.~Casella.
\newblock {Theory of Point Estimation, Second Edition}.
\newblock Springer Science \& Business Media, pp. 125, 1998.

\bibitem{levy}
A.~Levy and M.~Lindenbaum.
\newblock {Sequential Karhunen–Loeve Basis Extraction and its Application to
  Images}.
\newblock \url{http://www.cs.technion.ac.il/~mic/doc/skl-ip.pdf} IEEE
  TRANSACTIONS ON IMAGE PROCESSING, VOL. 9, NO. 8, 2000.

\bibitem{litjens}
G.~Litjens, T.~Kooi, B.~E. Bejnordi, A.~A.~A. Setio, F.~Ciompi, M.~Ghafoorian,
  J.~A. W.~M. van~der Laak, B.~van Ginneken, and C.~I. Sánchez.
\newblock A survey on deep learning in medical image analysis.
\newblock
  \url{http://www.sciencedirect.com/science/article/pii/S1361841517301135}
  Medical Image Analysis, vol. 42, pp. 60-88, 2017.

\bibitem{mackay}
D.~MacKay.
\newblock {A practical Bayesian framework for backpropagation networks}.
\newblock \url{http://www.inference.org.uk/mackay/PhD.html#PhD}, Neural
  Computation, 4(3):448–472, 1992., 1992.

\bibitem{martens}
J.~Martens.
\newblock {New insights and perspectives on the natural gradient method}.
\newblock \url{https://arxiv.org/abs/1412.1193} arXiv:1412.1193 [cs.LG], 2020.

\bibitem{murfet}
D.~Murfet, S.~Wei, M.~Gong, H.~Li, J.~Gell-Redman, and T.~Quella.
\newblock {Deep Learning is Singular, and That's Good}.
\newblock \url{https://arxiv.org/abs/2010.11560} arXiv:2010.11560 [cs.LG],
  2020.

\bibitem{nagarajan}
P.~Nagarajan and G.~Warnell.
\newblock {Deterministic Implementations for Reproducibility in Deep
  Reinforcement Learning}.
\newblock \url{https://arxiv.org/abs/1809.05676}, arXiv:1809.05676 [cs.AI],
  2019.

\bibitem{mcfadden}
W.~K. Newey and D.~McFadden.
\newblock {Handbook of Econometrics}.
\newblock
  \url{https://www.sciencedirect.com/science/article/pii/S1573441205800054},
  1994.

\bibitem{nilsen}
G.~K. Nilsen, A.~Z. Munthe-Kaas, H.~J. Skaug, and M.~Brun.
\newblock {Efficient Computation of Hessian Matrices in TensorFlow}.
\newblock \url{https://arxiv.org/abs/1905.05559}, arXiv:1905.05559v1 [cs.LG],
  2019.

\bibitem{osband}
I.~Osband.
\newblock {Risk versus Uncertainty in Deep Learning: Bayes, Bootstrap and the
  Dangers of Dropout}.
\newblock \url{http://bayesiandeeplearning.org/2016/papers/BDL_4.pdf}, NIPS
  Workshop on Bayesian Deep Learning, 2016.

\bibitem{osband2}
I.~Osband, C.~Blundell, A.~Pritzel, and B.~V. Roy.
\newblock {Deep Exploration via Bootstrapped DQN}.
\newblock
  \url{https://papers.nips.cc/paper/6501-deep-exploration-via-bootstrapped-dqn.pdf},
  Conference on Neural Information Processing Systems (NIPS), 2016.

\bibitem{pearlmutter}
B.~A. Pearlmutter.
\newblock {Fast Exact Multiplication by the Hessian}.
\newblock \url{http://www.bcl.hamilton.ie/~barak/papers/nc-hessian.pdf}, Neural
  Computation, 6 (1) (1994), pp. 147-160, 1994.

\bibitem{sagun}
L.~Sagun, L.~Bottou, and Y.~LeCun.
\newblock Eigenvalues of the hessian in deep learning: Singularity and beyond.
\newblock \url{https://arxiv.org/abs/1611.07476}, arXiv:1611.07476v2 [cs.LG],
  2017.

\bibitem{sagun2}
L.~Sagun, U.~Evci, V.~U. Guney, Y.~Dauphin, and L.~Bottou.
\newblock Empirical analysis of the hessian of over-parametrized neural
  networks.
\newblock \url{https://arxiv.org/abs/1706.04454}, arXiv:1706.04454v3 [cs.LG],
  2018.

\bibitem{schulam}
P.~Schulam and S.~Saria.
\newblock {Can You Trust This Prediction? Auditing Pointwise Reliability After
  Learning}.
\newblock \url{https://arxiv.org/abs/1901.00403} arXiv:1901.00403 [stat.ML],
  2019.

\bibitem{song}
H.~Song, M.~Kim, D.~Park, and J.-G. Lee.
\newblock {Learning from Noisy Labels with Deep Neural Networks: A Survey}.
\newblock \url{https://arxiv.org/pdf/2007.08199}, arXiv:2007.08199v2 [cs.LG],
  2020.

\bibitem{pydeepdeltamodule}
{pyDeepDelta: A TensorFlow Module Implementing the Delta Method in Deep
  Learning Classification}.
\newblock \url{https://github.com/gknilsen/pydeepdelta.git}.

\bibitem{sklearn}
scikit-learn.
\newblock \url{https://scikit-learn.org/}.

\bibitem{scipy}
Scipy.
\newblock \url{http://www.scipy.org}.

\bibitem{trefethen}
L.~N. Trefethen and D.~B. III.
\newblock {Numerical Linear Algebra}.
\newblock Siam, 1997.

\bibitem{watanabe}
S.~Watanabe.
\newblock {Almost All Learning Machines are Singular}.
\newblock
  \url{http://watanabe-www.math.dis.titech.ac.jp/users/swatanab/foci2007.pdf},
  Proceedings of the 2007 IEEE Symposium on Foundations of Computational
  Intelligence, 2007.

\bibitem{zhu}
L.~Zhu and N.~Laptev.
\newblock {Deep and Confident Prediction for Time Series at Uber}.
\newblock 2017 IEEE International Conference on Data Mining Workshops (ICDMW),
  2017.

\end{thebibliography}

\end{document}